\documentclass{article} % For LaTeX2e
\usepackage{iclr2017_conference,times}
\usepackage{hyperref}
\usepackage{url}

\usepackage[utf8]{inputenc} % allow utf-8 input
\usepackage[T1]{fontenc}    % use 8-bit T1 fonts

\usepackage{booktabs}
\usepackage{amsmath,amsfonts}
\usepackage{graphicx}
\usepackage{xspace,color}
\usepackage{bbm}
\usepackage{multirow}
\usepackage{subcaption}

\title{Improving Neural Language Models with a Continuous Cache}

\author{Edouard Grave, Armand Joulin, Nicolas Usunier \\
        Facebook AI Research \\
        \texttt{\{egrave,ajoulin,usunier\}@fb.com}}

\usepackage{tikz}
\usetikzlibrary{positioning}
\usetikzlibrary{shapes.misc}
\begin{document}

\maketitle

\begin{abstract}
We propose an extension to neural network language models to
adapt their prediction to the recent history. Our model
is a simplified version of memory augmented networks, which stores
past hidden activations as memory and accesses them through a dot product
with the current hidden activation. This mechanism is very efficient
and scales to very large memory sizes. We also draw a link between
the use of external memory in neural network and cache models used
with count based language models. We demonstrate on several language
model datasets that our approach performs significantly better than
recent memory augmented networks.
\end{abstract}

\section{Introduction}

Language modeling is a core problem in natural language
processing, with many applications such as machine
translation~\citep{brown1993mathematics}, speech
recognition~\citep{bahl1983maximum} or dialogue
agents~\citep{stolcke2000dialogue}.  While traditional neural networks language
models have obtained state-of-the-art performance in this
domain~\citep{jozefowicz2016exploring,mikolov2010recurrent}, they lack
the capacity to adapt to their recent history, limiting their application to
dynamic environments~\citep{dodge2015evaluating}. A recent
approach to solve this problem is to augment these networks with an external
memory~\citep{graves2014neural,grefenstette2015learning,joulin2015inferring,sukhbaatar2015end}.
These models can potentially use their external memory to store new
information and adapt to a changing environment.

While these networks have obtained promising results on
language modeling datasets~\citep{sukhbaatar2015end}, they are quite
computationally expensive. Typically, they have to learn a parametrizable
mechanism to read or write to memory cells ~\citep{graves2014neural,joulin2015inferring}.
This may limit both the size of their usable memory 
as well as the quantity of data they can be trained on. In this work, we
propose a very light-weight alternative that shares some of the properties of
memory augmented networks, notably the capability to dynamically adapt over time.
By minimizing the computation burden of the memory, we are able to use larger
memory and scale to bigger datasets. We observe in practice that this allows
us to surpass the perfomance of memory augmented networks on different language
modeling tasks.

Our model share some similarities with a model proposed
by~\citet{kuhn1988speech}, called the \emph{cache model}. A cache model stores a
simple representation of the recent past, often in the form of unigrams, and uses
them for prediction~\citep{kuhn1990cache}. This contextual information
is quite cheap to store and can be accessed efficiently. It also does not need
any training and can be appplied on top of any model. This makes this model particularly
interesting for domain adaptation~\citep{kneser1993dynamic}.

Our main contribution is to propose a continuous version of the cache model, called \emph{Neural Cache Model},
that can be adapted to any neural network language model. We store recent hidden activations and use
them as representation for the context. Using simply a dot-product with the current hidden activations, they
turn out to be extremely informative for prediction. Our model requires \emph{no training} and can be used
on any pre-trained neural networks. It also scales effortlessly to thousands of memory cells.
We demonstrate the quality of the Neural Cache models on several language model tasks and the LAMBADA dataset~\citep{paperno2016lambada}.

\section{Language modeling}

A language model is a probability distribution over sequences of words.
Let $V$ be the size of the vocabulary; each word is represented by a one-hot
encoding vector $x$ in $\mathbb{R}^V = \mathcal{V}$, corresponding to its index in
the vocabulary.
Using the chain rule, the probability assigned to a sequence of words $x_1,\dots, x_T$ can be factorized as
\begin{equation*}
p(x_1, ..., x_T) = \prod_{t=1}^T p(x_t \ | \ x_{t-1}, ..., x_1).
\end{equation*}
Language modeling is often framed as learning the conditional probability over words, given the history~\citep{bahl1983maximum}.

This conditional probability is traditionally approximated with non-parameteric models based on counting statistics~\citep{goodman2001bit}.
In particular, smoothed N-gram
models~\citep{katz1987estimation,kneser1995improved} achieve good performance
in practice~\citep{mikolov2011empirical}. Parametrized alternatives are either
maximum entropy language models~\citep{rosenfeld1996maximum},
feedforward networks~\citep{bengio2003neural} or  recurrent
networks~\citep{mikolov2010recurrent}. In particular, recurrent networks are
currently the best solution to approximate this conditional probability,
achieving state-of-the-arts performance on standard language
modeling benchmarks~\citep{jozefowicz2016exploring,zilly2016recurrent}.

\paragraph{Recurrent networks.}
Assuming that we have a vector $h_t \in \mathbb{R}^d$ encoding the history~$x_t, ..., x_1$, the conditional probability of 
a word $w$ can be parametrized as
\begin{equation*}
p_{vocab}( w \ | \ x_t, ..., x_1) \propto \exp(h_t^{\top} o_w).
\end{equation*}
The history vector $h_t$ is computed by a recurrent network by recursively applying 
an equation of the form
\begin{equation*}
h_t = \Phi\left(x_t, h_{t-1} \right),
\end{equation*}
where $\Phi$ is a function depending on the architecture of the network.
Several architecture for recurrent networks have been proposed, such as the
Elman network~\citep{elman1990finding}, the long short-term memory~
(LSTM)~\citep{hochreiter1997long} or the gated recurrent
unit~(GRU)~\citep{chung2014empirical}.
One of the simplest recurrent networks is the Elman network~\citep{elman1990finding}, where
\begin{equation*}
h_t = \sigma \left( L x_t + R h_{t-1} \right),
\end{equation*}
where $\sigma$ is a non-linearity such as the logistic or tanh functions, $L
\in \mathbb{R}^{d \times V}$ is a word embedding matrix and $R \in
\mathbb{R}^{d \times d}$ is the recurrent matrix.
The LSTM architecture is particularly interesting in the context of language modelling~\citep{jozefowicz2016exploring}
and we refer the reader to~\cite{graves2013speech} for details on this architecture. 

The parameters of recurrent neural network language models are learned by
minimizing the negative log-likelihood of the training data.  This objective
function is usually minimized by using the stochastic gradient descent
algorithm, or variants such as Adagrad~\citep{duchi2011adaptive}.  The gradient
is computed using the truncated backpropagation through time
algorithm~\citep{werbos1990backpropagation,williams1990efficient}.

\paragraph{Cache model.}
After a word appears once in a document, it is much more likely to appear again.
As an example, the frequency of the word \emph{tiger} on the Wikipedia page of the same name is $2.8\%$, compared to $0.0037\%$ over the whole Wikipedia.
Cache models exploit this simple observation to improve $n$-gram language models by capturing long-range dependencies in documents.
More precisely, these models have a cache component, which contains the words that appeared in the recent history (either the document or a fixed number of words).
A simple language model, such as a unigram or smoothed bigram model, is fitted on the words of the cache and interpolated with the static language model (trained over a larger dataset).
This technique has many advantages.
First, this is a very efficient way to adapt a language model to a new domain.
Second, such models can predict out-of-vocabulary words (OOV words), after seeing them once.
Finally, this helps capture long-range dependencies in documents, in order to generate more coherent text.

\newpage
\section{Neural Cache Model}

\begin{figure}[t]
\centering
\begin{minipage}{0.55\linewidth}
\begin{tikzpicture}
    \tikzstyle{hidden}=[draw=black,minimum height=30pt]
    \tikzstyle{cell}=[draw=black,minimum height=15pt,minimum width=43pt]
    %    \tikzstyle{arrow}=[-{Latex[scale=1.0]},draw=black]
    \tikzstyle{arrow}=[-latex,draw=black]

    \node (h0) at (0,   0) [hidden] {$h_1$};
    \node (h1) at (2, 0) [hidden] {$h_2$};
    \node (h2) at (4,   0) [hidden] {$h_3$};
    \node (h3) at (6, 0) [hidden] {$h_4$};

    \node (x0) at (0, -2) {$x_1$};
    \node (x1) at (2, -2) {$x_2$};
    \node (x2) at (4, -2) {$x_3$};
    \node (x3) at (6, -2) {$x_4$};

    \node (y1) at (.5, 2) [cell] {$(h_1,x_2)$};
    \node (y2) at (2, 2) [cell] {$(h_2,x_3)$};
    \node (y3) at (3.5, 2) [cell] {$(h_3,x_4)$};
    \node (y4) at (6, 2) {$x_5$};

    \draw[arrow] (x0.north) -- node[right] {$L$} (h0.south);
    \draw[arrow] (x1.north) -- node[right] {$L$} (h1.south);
    \draw[arrow] (x2.north) -- node[right] {$L$} (h2.south);
    \draw[arrow] (x3.north) -- node[right] {$L$} (h3.south);

    \draw[arrow] (h0.east) -- node[above] {$R$} (h1.west);
    \draw[arrow] (h1.east) -- node[above] {$R$} (h2.west);
    \draw[arrow] (h2.east) -- node[above] {$R$} (h3.west);

    \draw[arrow] (h0.north) -- node[right] {$Id$} (y1.south);
    \draw[arrow] (h1.north) -- node[right] {$Id$} (y2.south);
    \draw[arrow] (h2.north) -- node[right] {$Id$} (y3.south);
    \draw[arrow] (h3.north) -- node[right] {$O$} (y4.south);

    \draw[arrow] (y3.east) -- node[above] {$Id$} (y4.west);
  \end{tikzpicture}
\end{minipage}
\begin{minipage}{0.38\linewidth}
\caption{The neural cache stores the previous hidden states
in memory cells. They are then used as keys to retrieve their
corresponding word, that is the next word. There is no
transformation applied to the storage during writing and reading.}
\label{fig:model}
\end{minipage}
\end{figure}
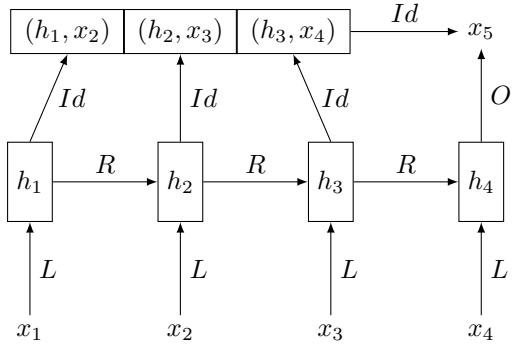

The Neural Cache Model adds a cache-like memory to neural network
language models.
It exploits the hidden representations $h_t$ to define a probability distribution over the words in the cache.
As illustrated Figure~\ref{fig:model}, the cache stores pairs $(h_i, x_{i+1})$ of a hidden
representation, and the word which was generated based on this representation
(we remind the reader that the vector $h_i$ encodes the history $x_i, ..., x_1$).
At time $t$, we then define a probability distribution over words stored in the cache based on the
stored hidden representations and the current one $h_t$ as
\begin{equation*}
p_{cache}( w \ | \ h_{1..t}, \ x_{1..t}) \propto \sum_{i=1}^{t-1} \mathbbm{1}_{\left\{ w = x_{i+1} \right\}} \exp (\theta h_t^{\top} h_i)
\end{equation*}
where the scalar $\theta$ is a parameter which controls the flatness of the distribution.
 When $\theta$ is equal to zero, the probability distribution over the history
 is uniform, and our model is equivalent to a unigram cache
 model~\citep{kuhn1990cache}.

From the point of view of memory-augmented neural networks, the
probability $p_{cache}( w \ | \ h_{1..t}, \ x_{1..t})$ given by the
neural cache model can be interpreted as the probability to retrieve
the word $w$ from the memory given the query $h_t$, where the desired
answer is the next word $x_{t+1}$. Using previous hidden states as
keys for the words in the memory, the memory lookup operator can be
implemented with simple dot products between the keys and the
query. In contrast to existing memory-augmented neural networks, the
neural cache model avoids the need to learn the memory lookup
operator. Such a cache can thus be added to a pre-trained recurrent
neural language model without fine tuning of the parameters, and large
cache size can be used with negligible impact on the computational
cost of a prediction.

\paragraph{Neural cache language model.}
Following the standard practice in n-gram cache-based language models,
the final probability of a word is given by  the linear interpolation of
the cache language model with the regular language model, obtaining:
\begin{equation*}
p(w \ | \ h_{1..t}, \ x_{1..t}) = (1 - \lambda) p_{vocab}(w \ | \ h_t) + \lambda p_{cache}(w \ | \ h_{1..t}, x_{1..t})\,.
\end{equation*}

Instead of taking a linear interpolation between the two distribution
with a fixed $\lambda$, we also consider a global normalization over
the two distribution:
\begin{equation*}
p( w \ | \ h_{1..t}, \ x_{1..t}) \propto \left( \exp(h_t^{\top} o_w) + \sum_{i=1}^{t-1} \mathbbm{1}_{\left\{ w = x_{i+1} \right\}} \exp (\theta h_t^{\top} h_i + \alpha) \right)\,.
\end{equation*}
This corresponds to taking a softmax over the vocabulary and the words in the cache.
The parameter $\alpha$ controls the weight of the cache component, and is the counterpart of the $\lambda$ parameter for linear interpolation.

The addition of the neural cache to a recurrent neural language model
inherits the advantages of $n$-gram caches in usual cache-based
models: The probability distribution over words is updated online
depending on the context, and out-of-vocabulary words can be predicted
as soon as they have been seen at least once in the recent
history. The neural cache also inherits the ability of the hidden
states of recurrent neural networks to model longer-term contexts than
small $n$-grams, and thus allows for a finer modeling of the
current context than e.g., unigram caches.

\paragraph{Training procedure.}
For now, we first train the (recurrent) neural network language model,
without the cache component.  We only apply the cache model at test
time, and choose the hyperparameters $\theta$ and $\lambda$ (or
$\alpha$) on the validation set. A big advantage of our
method is that it is very easy and cheap to apply, with
already trained neural models.  There is no need to perform
backpropagation over large contexts, and we can thus apply our method
with large cache sizes (larger than one thousand).

\section{Related work}

\paragraph{Cache model.}
Adding a cache to a language model was intoducted in the context of speech
recognition\citep{kuhn1988speech,kupiec1989probabilistic,kuhn1990cache}.
These models were further extended by \citet{jelinek1991dynamic} into a smoothed
trigram language model, reporting reduction in both perplexity and word error rates.
\citet{della1992adaptive} adapt the cache to a general $n$-gram model
such that it satisfies marginal constraints obtained from the current document.

\paragraph{Adaptive language models.} Other adaptive language models have been
proposed in the past: \citet{kneser1993dynamic} and \citet{iyer1999modeling}
dynamically adapt the parameters of their model to the recent history using
different weight interpolation schemes.  \citet{bellegarda2000exploiting} and
\citet{coccaro1998towards} use latent semantic analysis to adapt their models
to the current context.  Similarly, topic features have been used with either
maximum entropy models~\citep{khudanpur2000maximum} or recurrent
networks~\citep{mikolov2012context,wang2015larger}.  Finally,
  \citet{lau1993trigger} proposes to use pairs of distant of words to capture
  long-range dependencies.

\begin{figure}
  \centering
  \includegraphics{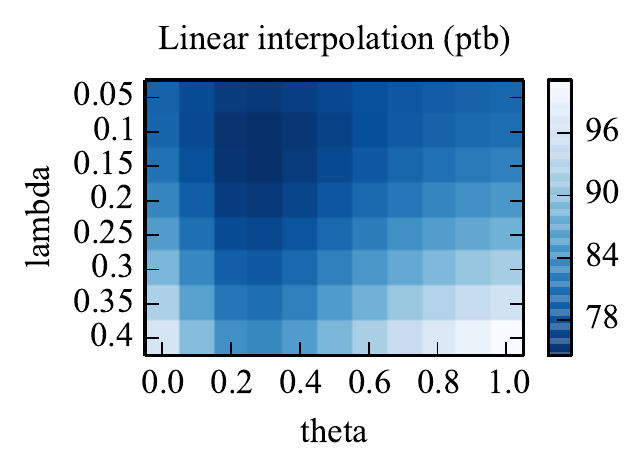}
  \includegraphics{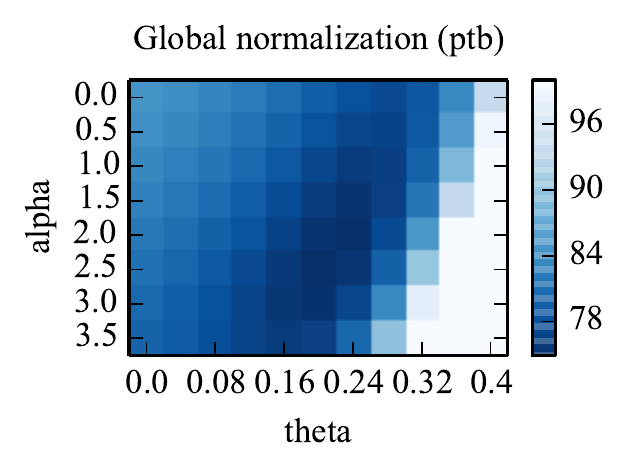}

  \vspace{-0.5cm}
  \caption{Perplexity on the validation set of \texttt{Penn Tree Bank} for linear interpolation (left) and global normalization (right), for various values of hyperparameters $\theta$, $\lambda$ and $\alpha$.
    We use a cache model of size $500$. The base model has a validation perplexity of $86.9$.
    The best linear interpolation has a perplexity of $74.6$, while the best global normalization has a perplexity of $74.9$.}
  \label{fig:ptb}
\end{figure}

\begin{table}
  \small
  \centering
  \begin{tabular}{lc}
    \toprule
    Model & Test PPL \\
    \midrule
    RNN+LSA+KN5+cache~\citep{mikolov2012context}         & 90.3 \\
    LSTM~\citep{zaremba2014recurrent}                    & 78.4 \\
    Variational LSTM~\citep{gal2015theoretically}        & 73.4 \\
    Recurrent Highway Network~\citep{zilly2016recurrent} & 66.0 \\ % 71.3
    Pointer Sentinel LSTM~\citep{merity2016pointer}      & 70.9 \\
    \midrule
    LSTM (our implem.)                                   & 82.3 \\
    Neural cache model                                   & 72.1 \\
    \bottomrule
  \end{tabular}
  \caption{Test perplexity on the \texttt{Penn Tree Bank}.}
  \label{tab:ptb}
\end{table}

\paragraph{Memory augmented neural networks.} 
In the context of sequence prediction, 
several memory augmented neural networks have obtained promising
results~\citep{sukhbaatar2015end,graves2014neural,grefenstette2015learning,joulin2015inferring}.
In particular,~\citet{sukhbaatar2015end} stores a representation of the 
recent past and accesses it using an attention mechanism~\cite{bahdanau2014neural}.
\citet{sukhbaatar2015end} shows that this reduces the perplexity for language
modeling.  This approach has been successfully applied to question
answering, when the answer is contained in a given
paragraph~\citep{chen2016thorough,hermann2015teaching,kadlec2016text,sukhbaatar2015end}.
Similarly, \citet{vinyals2015pointer} explores the use of this mechanism to
reorder sequences of tokens. Their network uses an attention (or ``pointer'')
over the input sequence to predict which element should be selected as the next
output.  \citet{gulcehre2016pointing} have shown that a similar mechanism
called \emph{pointer softmax} could be used in the context of machine
translation,  to decide which word to copy from the source to target.  

Independently of our work, \citet{merity2016pointer} apply the same mechanism
to recurrent network. Unlike our work, they uses the current hidden activation
as a representation of the current input (while we use it to represent the
    output). This requires additional learning of a transformation between the current
representation and those in the past. The advantage of our
approach is that we can scale to very large caches effortlessly.

\section{Experiments}

In this section, we evaluate our method on various language modeling datasets, which have different sizes and characteristics.
On all datasets, we train a static recurrent neural network language model with LSTM units.
We then use the hidden representations from this model to obtain our cache, which is interpolated with the static LSTM model.
We also evaluate a unigram cache model interpolated with the static model as another baseline.

\begin{figure}
  \centering
  \includegraphics{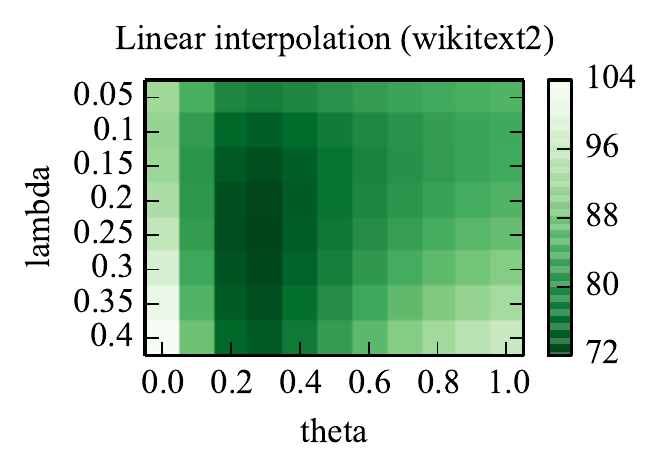}
  \includegraphics{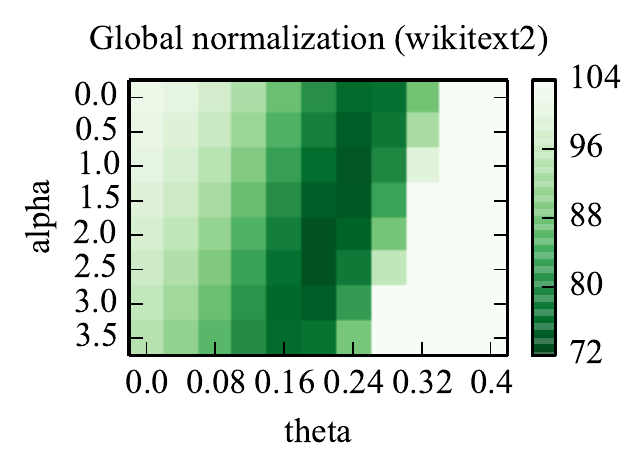}

  \vspace{-0.5cm}
  \caption{Perplexity on the validation set of \texttt{wikitext2} for linear interpolation (left) and global normalization (right), for various values of hyperparameters $\theta$, $\lambda$ and $\alpha$.
    We use a cache model of size $2000$. The base model has a validation perplexity of $104.2$.
    The best linear interpolation has a perplexity of $72.1$, while the best global normalization has a perplexity of $73.5$.}
  \label{fig:wikitext2}
\end{figure}

\begin{table}
  \small
  \centering
  \begin{tabular}{lcc}
    \toprule
    Model & \texttt{wikitext2} & \texttt{wikitext103} \\
    \midrule
    Zoneout + Variational LSTM~\citep{merity2016pointer}  & 100.9 & - \\
    Pointer Sentinel LSTM~\citep{merity2016pointer}       & 80.8  & - \\
    \midrule
    LSTM (our implementation)         & 99.3 & 48.7 \\
    Neural cache model (size = 100)   & 81.6 & 44.8 \\
    Neural cache model (size = 2,000) & 68.9 & 40.8 \\
    \bottomrule
  \end{tabular}
  \caption{Test perplexity on the \texttt{wikitext} datasets. The two datasets share the same validation and test sets, making all the results comparable.}
  \label{tab:wikitext}
\end{table}

\subsection{Small scale experiments}
\paragraph{Datasets.} In this section, we describe experiments performed on two small datasets: the \texttt{Penn Tree Bank}~\citep{marcus1993building} and the \texttt{wikitext2}~\citep{merity2016pointer} datasets.
The \texttt{Penn Tree Bank} dataset is made of articles from the Wall Street Journal, contains 929k training tokens and has a vocabulary size of 10k.
The \texttt{wikitext2} dataset is derived from Wikipedia articles, contains 2M training tokens and has a vocabulary size of 33k.
These datasets contain non-shuffled documents, therefore requiring models to capture inter-sentences dependencies to perform well.

\paragraph{Implementation details.} We train recurrent neural network language models with 1024 LSTM units, regularized with dropout (probability of dropping out units equals to $0.65$).
We use the Adagrad algorithm, with a learning rate of $0.2$, a batchsize of $20$ and initial weight uniformly sampled in the range~$[-0.05, 0.05]$.
We clip the norm of the gradient to $0.1$ and unroll the network for $30$ steps.
We consider cache sizes on a logarithmic scale, from $50$ to $10,000$, and fit the cache hyperparameters on the validation set.

\paragraph{Results.} We report the perplexity on the validation sets in Figures~\ref{fig:ptb} and \ref{fig:wikitext2}, for various values of hyperparameters, for linear interpolation and global normalization.
First, we observe that on both datasets, the linear interpolation method performs slightly better than the global normalization approach.
It is also easier to apply in practice, and we thus use this method in the remainder of this paper.
In Tables~\ref{tab:ptb} and \ref{tab:wikitext}, we report the test perplexity of our approach and state-of-the-art models.
Our approach is competitive with previous models, in particular with the pointer sentinel LSTM model of \citet{merity2016pointer}.
On \texttt{Penn Tree Bank}, we note that the improvement over the base model is similar for both methods.
On the \texttt{wikitext2} dataset, both methods obtain similar results when using the same cache size ($100$ words).
Since our method is computationally cheap, it is easy to increase the cache to larger values ($2,000$ words), leading to dramatic improvements~($30\%$ over the baseline, $12\%$ over a small cache of $100$ words).

\subsection{Medium scale experiments}
\paragraph{Datasets and implementation details.} In this section, we describe experiments performed over two medium scale datasets: \texttt{text8} and \texttt{wikitext103}.
Both datasets are derived from Wikipedia, but different pre-processing were applied.
The \texttt{text8} dataset contains 17M training tokens and has a vocabulary size of 44k words, while the \texttt{wikitext103} dataset has a training set of size 103M, and a vocabulary size of 267k words.
We use the same setting as in the previous section, except for the batchsize (we use $128$) and dropout parameters (we use $0.45$ for \texttt{text8} and $0.25$ for \texttt{wikitext103}).
Since both datasets have large vocabularies, we use the adaptive softmax~\citep{grave2016efficient} for faster training.

\begin{figure}
  \centering
  \includegraphics{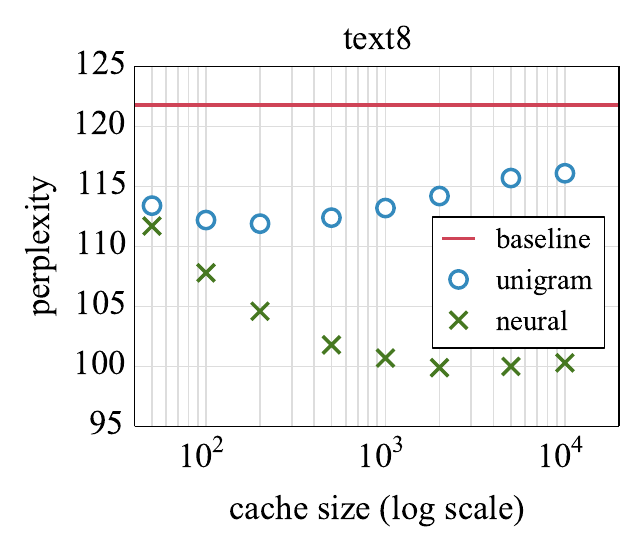}
  \hfill
  \includegraphics{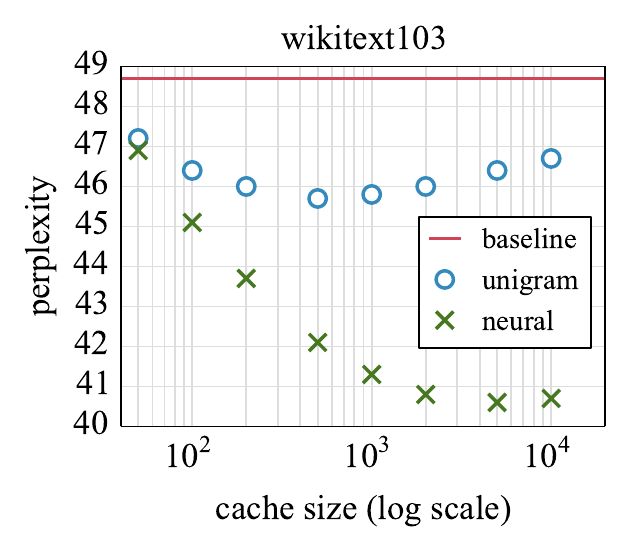}

  \vspace{-0.3cm}
  \caption{Test perplexity as a function of the number of words in the cache, for our method and a unigram cache baseline.
    We observe that our approach can uses larger caches than the baseline.}
  \label{fig:cachesize}
\end{figure}

\paragraph{Results.} We report the test perplexity as a function of the cache size in Figure~\ref{fig:cachesize}, for the neural cache model and a unigram cache baseline.
We observe that our approach can exploits larger cache sizes, compared to the baseline.
In Table~\ref{tab:wikitext}, we observe that the improvement in perplexity of our method over the LSTM baseline on \texttt{wikitext103} is smaller than for \texttt{wikitext2}~(approx. $16\%$ v.s. $30\%$).
The fact that improvements obtained with more advanced techniques decrease when the size of training data increases has already been observed by \citet{goodman2001bit}.
Both \texttt{wikitext} datasets sharing the same test set,  we also observe that the LSTM baseline, trained on 103M tokens (\texttt{wikitext103}), strongly outperforms more sophisticated methods, trained on 2M tokens (\texttt{wikitext2}).
For these two reasons, we believe that it is important to evaluate and compare methods on relatively large datasets.

\begin{table}
  \small
  \centering
  \begin{subtable}{0.44 \textwidth}
  \begin{tabular}{lc}
    \toprule
    Model & Test \\
    \midrule
    LSTM-500~\citep{mikolov2014learning}   & 156 \\
    SCRNN~\citep{mikolov2014learning}      & 161 \\
    MemNN~\citep{sukhbaatar2015end}        & 147 \\
    \midrule
    LSTM-1024 (our implem.)                & 121.8 \\
    Neural cache model                     &  99.9 \\
    \bottomrule
  \end{tabular}
  \caption{\texttt{text8}}
  \end{subtable}
  \hfill
  \begin{subtable}{0.50 \textwidth}
  \begin{tabular}{lcc}
    \toprule
    Model & Dev & Ctrl \\
    \midrule
    WB5~\citep{paperno2016lambada}       & 3125 & 285 \\
    WB5+cache~\citep{paperno2016lambada} &  768 & 270 \\
    LSTM-512~\citep{paperno2016lambada}           & 5357 & 149 \\
    \midrule
    LSTM-1024 (our implem.)                   & 4088 &  94 \\
    Neural cache model                        &  138 & 129 \\
    \bottomrule
  \end{tabular}
  \caption{\texttt{lambada}}
  \end{subtable}

  \caption{Perplexity on the \texttt{text8} and \texttt{lambada} datasets. WB5 stands for $5$-gram language model with Witten-Bell smoothing.}
  \label{tab:text8}
\end{table}

\subsection{Experiments on the lambada dataset}
\begin{figure}
  \centering
  \includegraphics{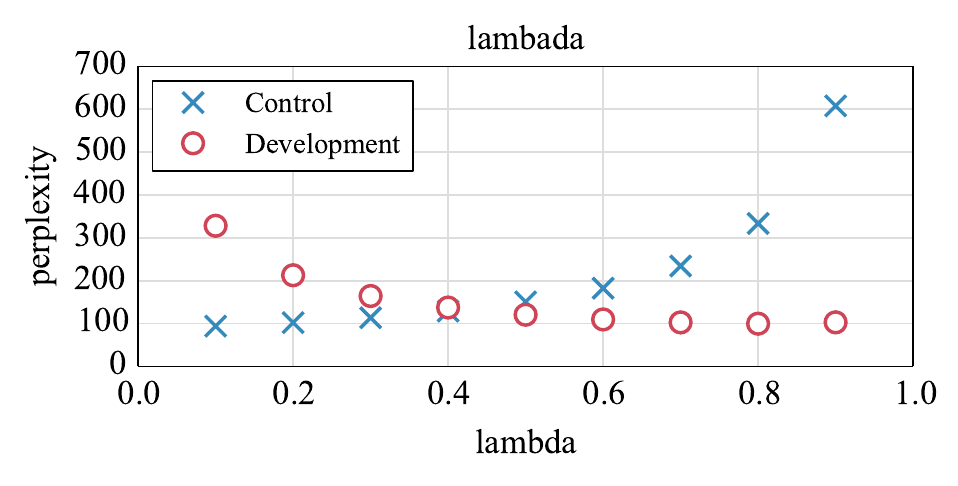}

  \vspace{-0.3cm}
  \caption{Perplexity on the development and control sets of \texttt{lambada}, as a function of the interpolation parameters $\lambda$.}
  \label{fig:lambada}
\end{figure}

Finally, we report experiments carried on the \texttt{lambada} dataset, introduced by \citet{paperno2016lambada}.
This is a dataset of short passages extracted from novels.
The goal is to predict the last word of the excerpt.
This dataset was built so that human subjects solve the task perfectly when given the full context (approx. 4.6 sentences), but fail to do so when only given the sentence with the target word.
Thus, most state-of-the-art language models fail on this dataset.
The \texttt{lambada} training set contains approximately 200M tokens and has a vocabulary size of $93,215$.
We report results for our method in Table~\ref{tab:text8}, as well the performance of baselines from \citet{paperno2016lambada}.
Adding a neural cache model to the LSTM baseline strongly improves the performance on the \texttt{lambada} dataset.
We also observe in Figure~\ref{fig:lambada} that the best interpolation parameter between the static model and the cache is not the same for the development and control sets.
This is due to the fact that more than 83\% of passages of the development set include the target word, while this is true for only 14\% of the control set.
Ideally, a model should have strong results on both sets.
One possible generalization of our model would be to adapt the interpolation parameter based on the current vector representation of the history $h_t$.

\section{Conclusion}

We presented the neural cache model to augment neural language models
with a longer-term memory that dynamically updates the word
probablilities based on the long-term context. A neural cache can be
added on top of a pre-trained language model at negligible cost. Our
experiments on both language modeling tasks and the challenging
LAMBADA dataset shows that significant performance gains can be
expected by adding this external memory component.

Technically, the neural cache models is similar to some recent
memory-augmented neural networks such as pointer networks. However,
its specific design makes it possible to avoid learning the memory
lookup component. This makes the neural cache appealing since
it can use larger cache sizes than memory-augment networks and can be
applied as easily as traditional count-based caches.

\small
\bibliography{iclr_cache}
\bibliographystyle{iclr2017_conference}

\end{document}